\documentclass[conference]{IEEEtran}
\IEEEoverridecommandlockouts
\usepackage{cite}
\usepackage{amsmath,amssymb,amsfonts}
\usepackage{algorithmic}
\usepackage{graphicx}
\usepackage{textcomp}
\usepackage{xcolor}
\usepackage[nohyperlinks,nolist]{acronym}
\usepackage{hyperref}
\def\BibTeX{{\rm B\kern-.05em{\sc i\kern-.025em b}\kern-.08em
    T\kern-.1667em\lower.7ex\hbox{E}\kern-.125emX}}
\begin{document}

\title{Augmented Reality-Enhanced Robot Teleoperation for Collecting User Demonstrations
\thanks{*This work was conducted at ABB Corporate Research as part of a Master's thesis project. Corresponding author: Zhen Li (e-mail: zhen.li@se.abb.com)}
}

\author{\IEEEauthorblockN{Shiqi Gong}
\IEEEauthorblockA{\textit{Aalto University}\\
Espoo, Finland \\
shiqi.gong@aalto.fi}
\and
\IEEEauthorblockN{Sebasti\'an Zudaire}
\IEEEauthorblockA{\textit{ABB Corporate Research}\\
Västerås, Sweden \\
sebastian.zudaire@se.abb.com}
\and
\IEEEauthorblockN{Chi Zhang}
\IEEEauthorblockA{\textit{ABB Corporate Research}\\
Västerås, Sweden \\
chi.zhang@se.abb.com}
\and
\IEEEauthorblockN{Zhen Li}
\IEEEauthorblockA{\textit{ABB Corporate Research}\\
Västerås, Sweden \\
zhen.li@se.abb.com}
}

\maketitle

\begin{abstract}
Traditional industrial robot programming is often complex and time-consuming, typically requiring weeks or even months of effort from expert programmers. Although Programming by Demonstration (PbD) offers a more accessible alternative, intuitive interfaces for robot control and demonstration collection remain challenging. To address this, we propose an Augmented Reality (AR)-enhanced robot teleoperation system that integrates AR-based control with spatial point cloud rendering, enabling intuitive, contact-free demonstrations. This approach allows operators to control robots remotely without entering the workspace or using conventional tools like the teach pendant. The proposed system is generally applicable and has been demonstrated on ABB robot platforms, specifically validated with the IRB 1200 industrial robot and the GoFa 5 collaborative robot. A user study evaluates the impact of real-time environmental perception, specifically with and without point cloud rendering, on task completion accuracy, efficiency, and user confidence. Results indicate that enhanced perception significantly improves task performance by 28\% and enhances user experience, as reflected by a 12\% increase in the System Usability Scale (SUS) score. This work contributes to the advancement of intuitive robot teleoperation, AR interface design, environmental perception, and teleoperation safety mechanisms in industrial settings for demonstration collection. The collected demonstrations may serve as valuable training data for machine learning applications.
\end{abstract}


\begin{acronym}[Orocos]
    \acro{PbD}[PbD]{Programming by Demonstration}
    \acro{AR}[AR]{Augmented Reality}
    \acro{VR}[VR]{Virtual Reality}
    \acro{EGM}[EGM]{Externally Guided Motion}
    \acro{RWS}[RWS]{Robot Web Services}
    \acro{rmse}[RMSE]{Root Mean Square Error}
    \acro{HMD}[HMD]{Head-Mounted Display}
    \acro{API}[API]{Application Programming Interface}
    \acro{TCP}[TCP]{Tool Center Point}
    \acro{ROS}[ROS]{Robot Operating System}
    \acro{URDF}[URDF]{Unified Robot Description Format}
    \acro{UI}[UI]{User Interface}
    \acro{REST}[REST]{Representational State Transfer}
    \acro{API}[API]{Application Programming Interface}
    \acro{SUS}[SUS]{System Usability Scale}
    \acro{CI}[CI]{Confidence Interval}
    \acro{UDP}[UDP]{User Datagram Protocol}
\end{acronym}

\section{Introduction}

\begin{figure*}[tb]
\centering
\includegraphics[width=\linewidth]{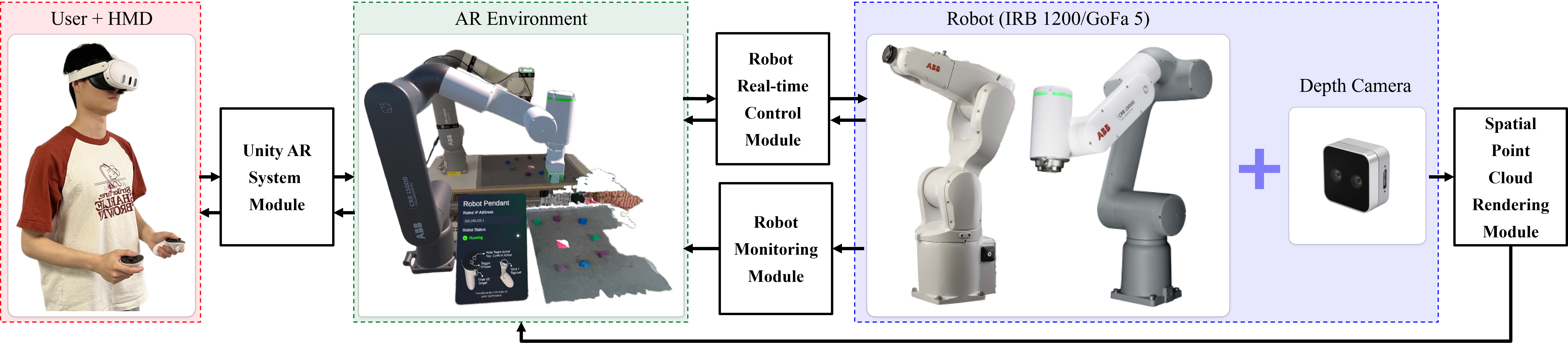}
\caption{System architecture of the AR-based robot teleoperation framework. The architecture comprises four core modules: (1) Robot Real-Time Control Module for low-latency communication via ABB’s EGM interface; (2) Unity AR System Module for immersive visualization and user interaction; (3) Spatial Point Cloud Rendering Module using Intel RealSense depth sensors for real-time environmental perception; and (4) Robot Monitoring Module utilizing ABB’s RWS for safety monitoring and feedback.}
\label{fig:system_architecture}
\end{figure*}

Industrial robots are increasingly deployed across manufacturing, logistics, and assembly domains, playing a central role in automation and productivity enhancement. However, traditional industrial robot programming is often complex, rigid and time-consuming, typically requiring weeks or even months of effort from skilled programmers \cite{ragaglia_accurate_2016,dean-leon_integration_2018}.

With growing demands for intelligent manufacturing and flexible small-batch production, industrial automation is shifting toward human-centered workflows. As robots transition from tools to collaborators, robot motion control, traditionally performed through teach pendants or scripted programming, is gradually moving toward more natural and intuitive interaction modalities \cite{lee_survey_2013}.

In this context, \ac{PbD} has emerged as a promising method, enabling robots to learn and replicate tasks through human demonstrations \cite{zhou2020advanced,jia_program_2016}. \ac{PbD} makes robot programming more accessible to non-experts and contributes to the development of intelligent robotic systems. Despite its potential, challenges persist in designing intuitive interfaces for robot control and demonstration collection.

A popular method for demonstrating a task in \ac{PbD} is lead-through teaching, where operators physically guide the robot along a trajectory. This method significantly lowers the learning curve by removing the need to learn programming languages or robot-specific control logic \cite{kirchner_intuitive_2015,geravand_human-robot_2013,liu_remote_2020}. However, this approach is typically limited to collaborative robots, which are equipped with specialized hardware such as torque sensors and compliant joints. These robots often suffer from high costs, limited payloads, and reduced speed, which limit their suitability in industrial settings demanding high performance  \cite{ragaglia_accurate_2016,zhang_sensorless_2019}. Additionally, traditional industrial robots still rely heavily on safety enclosures, making physical interaction both difficult and risky \cite{matheson_humanrobot_2019}.

This work proposes an \ac{AR}-enhanced teleoperation system that enables virtual lead-through teaching for industrial robots lacking native lead-through functionality. The system is demonstrated on multiple ABB platforms, including the IRB 1200 industrial robot and the GoFa 5 collaborative robot, enabling consistent, intuitive interaction across different hardware types. By integrating real-time point cloud rendering, users can interact with both virtual and physical robots in a contact-free manner, significantly improving visual feedback and task awareness. This approach aims to reduce demonstration complexity, shorten deployment cycles, and support safer, more flexible human-robot collaboration in industrial contexts.

The main contributions of this AR-enhanced teleoperation system are as follows:
\begin{itemize}
    \item designed an \ac{AR}-based teleoperation system for industrial robots without lead-through functionality, with compatibility extended to collaborative robots that natively support lead-through operations;
    \item integrated intuitive \ac{AR}-based teleoperation control with real-time spatial point cloud rendering for enhanced visual feedback and contact-free demonstrations;
    \item conducted a user study to evaluate the impact of real-time environmental perception, involving five participants with diverse levels of experience in \ac{AR}/\ac{VR} and robotics.
\end{itemize}

\section{Related Work}\label{sec:literature}

Several non-contact teleoperation techniques have been developed for demonstrating a task. These include vision-based motion tracking using markers \cite{ferreira_stereo-based_2016}, audio-visual fusion systems integrating speech and gestures \cite{du_online_2018}, and wearable sensor-based interfaces \cite{javaid_using_2015}. While these methods reduce safety risks and eliminate the need for direct physical contact, they often suffer from drawbacks such as high setup costs, limited adaptability to dynamic environments, insufficient intuitiveness, and challenges in achieving precise, real-time control. Moreover, a common issue across many of these methods is the motion transfer gap between human input and robot execution, which refers to the difficulty in accurately and seamlessly mapping human motions to robotic actions. These challenges limit the effectiveness of existing non-contact teleoperation approaches in real-world industrial applications.

The development of \ac{AR} and \ac{VR} offers new opportunities for non-contact, intuitive robot interaction \cite{de_pace_systematic_2020,chandan2021arroch,kennel2023interacting}. By overlaying virtual content onto the physical world, \ac{AR} enhances spatial perception and interaction fidelity in real-time 3D environments \cite{zhang_see_2023,zhang_supporting_2021,zhang_playing_2023}. AR-based interfaces can improve the efficiency and accessibility of robot programming and teleoperation, while also reducing reliance on costly hardware like torque sensors \cite{othman_humanrobot_2023}.

Rivera-Pinto et al. \cite{rivera2024toward} proposed an \ac{AR} interface that allows users to define the trajectory of a robot end-effector by placing discrete holographic spheres in the workspace. A similar concept is explored in \cite{quintero_robot_2018}. However, the feasibility of the defined trajectories is not guaranteed, as the reachability of these discrete points by the robot is not validated at the time of definition. Thormann et al. \cite{thormann_gesture_2023} introduced an \ac{AR} interface for commanding robot actions using predefined hand gestures. While intuitive, this approach imposes limitations on the robot's motion capabilities and lacks support for interaction in dynamic environments. Smith et al. \cite{smith_augmented_2025} and Van Haastregt et al. \cite{van2024puppeteer} presented an \ac{AR} interface for robot teleoperation, enabling users to control a physical robot via a digital end effector surrogate. However, users must rely on direct visual observation of the physical environment to manipulate objects interacting with the robot, which constrains both the allowable distance from the robot and the size of the objects that can be effectively handled due to visibility limitations. Pizzagalli et al. \cite{pizzagalli_ros-based_2024} proposed an \ac{AR} path planning interface that enables users to define robot trajectories within a digital twin environment. However, this approach requires a replicated model of the physical environment in the virtual setup, which limits its adaptability to dynamic or changing environments.

Our work extends the concept of AR-based teleoperation, where users directly manipulate a virtual robot and the physical robot mirrors the motion in real time, as demonstrated in \cite{van2024puppeteer, smith_augmented_2025}. Expanding upon this concept, we propose a modular \ac{AR} teleoperation system with real-time point cloud rendering to enhance spatial awareness and interaction precision.

\section{Method}\label{sec:method}

The proposed universal AR-based robot demonstration system consists of four core modules, which are relatively independent yet highly integrated in their operation:
\begin{itemize}
    \item \textit{Robot Real-time Control Module} that utilizes ABB’s \ac{EGM} interface to enable low-latency communication and precise robot control;
    \item \textit{Unity AR System Module} that provides immersive visualization and intuitive user interaction within the augmented environment;
    \item \textit{Spatial Point Cloud Rendering Module} that delivers real-time environmental perception through integration with Intel RealSense depth sensors;
    \item \textit{Robot Monitoring Module} that ensures operational safety through continuous feedback and status updates via ABB's \ac{RWS}.
\end{itemize}

The details of each module are described in the following subsections. An overview of the system architecture, including these four modules, is illustrated in Fig.~\ref{fig:system_architecture}.

The proposed system is built upon \ac{AR} technology and integrates key hardware components: \ac{HMD}, ABB industrial and collaborative robots (IRB 1200 and GoFa 5), and a depth sensor. The \ac{HMD} serves as the primary display interface, offering immersive \ac{AR} visualization along with spatial interaction and tracking capabilities. Robots act as the execution units, receiving motion commands and performing the corresponding actions. The depth sensor provides real-time 3D point cloud data for environmental perception.

The system follows a modular architecture centered around the Unity-based \ac{AR} environment for data processing and interaction. Users demonstrate robot motions in the \ac{AR} interface using the \ac{HMD}. Interaction data is processed by the Unity-based \ac{AR} module and transmitted between the virtual and physical robots through the Robot Real-Time Control Module. At the same time, robot status is streamed back through the Robot Monitoring Module for real-time feedback. A depth sensor captures point cloud data, enabling spatial awareness by rendering the robot’s surroundings in AR. An event-driven communication scheme ensures loose coupling between modules, improving scalability and maintainability.

\subsection{Robot Real-time Control Module}

The development of the Robot Real-Time Control Module focuses on two core aspects: selecting an effective communication strategy and optimizing motion control precision.

For ABB robots, \ac{EGM} offers distinct technical advantages, making it well-suited for responsive and low-latency control in AR-based teleoperation scenarios: (1) \ac{EGM} provides a standardized communication interface across various ABB robot models, simplifying the integration of multi-robot systems; (2) Native motion interpolation and planning eliminate the need for intermediate motion planning components; (3) \ac{EGM} establishes a direct, point-to-point connection with the robot controller via \ac{UDP}, supporting a control loop at 250 Hz. This architecture achieves end-to-end communication latency as low as 4 ms. Based on these advantages, \ac{EGM} is adopted as the core communication solution in this study.

\ac{EGM} supports three control modes: position stream, position guidance, and path correction. This work primarily employs the position stream mode, which allows external applications to continuously transmit target positions to the robot. The controller then performs real-time interpolation and motion planning to ensure smooth and safe execution.

System development followed an iterative design process, with particular emphasis during prototype validation on coordinate frame transformations between Unity’s left-handed system and ABB’s right-handed robot frame. Accurate transformation is essential for consistent mapping between the virtual and physical robots, supporting intuitive and real-time teleoperation. The transformation of the desired \ac{TCP} pose from Unity to the robot system involves both axis reordering and unit conversion from meters to millimeters.

To ensure continuous synchronization between the virtual and physical robots, the system streams the desired \ac{TCP} pose from the virtual robot to the real robot via \ac{EGM}. The virtual robot's joint states are then continuously updated based on the actual joint positions returned by the physical robot.

To replicate joint positions of the physical robot within Unity, a joint angle mapping mechanism is implemented. Experimental testing revealed that joint rotation directions in Unity are opposite to those transmitted via \ac{EGM}. To address this, a sign inversion is applied based on the axis of rotation:
\begin{equation}
\vec{v}_{joint} = -\theta_{EGM} \cdot \hat{e}
\end{equation}
where \( \hat{e} \in \{(1, 0, 0),\ (0, 1, 0),\ (0, 0, 1)\} \) corresponds to rotation about the X-, Y-, or Z-axis, respectively. $\vec{v}_{joint}$ represents the local Euler angle vector applied to the Unity joint transform, and $\theta_{EGM}$ denotes the joint angle received from EGM. This correction mechanism ensures that the motion of the virtual robot remains fully synchronized with the physical robot, providing the operator with accurate and consistent visual feedback.

To enhance motion control precision, this module integrates two key safety mechanisms: speed-limited following and anti-jitter filtering. Both are implemented in accordance with industrial safety standards and ABB’s technical guidelines.

Speed-limited following is configured using the $MaxSpeedDeviation$ parameter in EGM, which defines the maximum allowable velocity deviation of the end-effector in Cartesian space. If the commanded velocity exceeds this threshold, the \ac{EGM} controller automatically adjusts the speed to remain within safe limits. In this study, $MaxSpeedDeviation$ is set to 50 mm/s, ensuring both motion smoothness and operator safety.

Anti-jitter filtering is achieved by configuring the $LpFilter$ parameter in EGM. This parameter applies a digital low-pass filter to smooth incoming position and orientation data, effectively mitigating high-frequency disturbances caused by communication delays, sensor noise, or user hand jitter. A cutoff frequency of 100 Hz is used, offering a balance between noise suppression and responsive control.

\subsection{Unity AR System Module}

The Unity \ac{AR} System Module focuses on \ac{AR} interface design and standardized visualization of robot models to support consistent cross-platform deployment. While \ac{URDF} is commonly used in robotics for robot modeling, typically importing STL or DAE formats via \ac{URDF} importers, it poses several limitations in \ac{AR} applications. STL models, based on low-resolution triangle meshes, often produce visible artifacts on curved surfaces. Additionally, \ac{URDF}-imported models frequently exhibit inconsistencies in coordinate systems and joint rotation axes when visualized in Unity, leading to inaccuracies in both appearance and motion behavior.

To address these issues, a standardized robot model processing workflow is developed to ensure consistent rendering and interaction across various ABB robot platforms in \ac{AR} environments. This workflow includes: (1) exporting high-fidelity FBX models from ABB RobotStudio; (2) reconstructing joint hierarchies and correcting rotation axes in Blender; (3) optimizing geometry and unifying materials to reduce rendering load; (4) applying consistent coordinate system transformations during export; and (5) validating the models in Unity via real-time synchronization with physical robots using the \ac{EGM} interface.

Interaction modality selection is based on comparative evaluation of hand gestures and physical controllers. Although gesture-based input offers intuitive control, testing showed that handheld controllers provide superior stability and accuracy, particularly for high-precision tasks such as rotation and positioning. Gesture input suffers from spatial ambiguity and susceptibility to unintentional motion, especially due to lack of stable spatial references and user variability in arm posture. In contrast, controllers offer physical anchoring, reducing unintended orientation changes and improving operational consistency. Furthermore, gesture recognition remains limited by latency and classification reliability, making it less suitable for real-time industrial teleoperation.

\begin{figure}[tb]
\centering
\includegraphics[width=\linewidth]{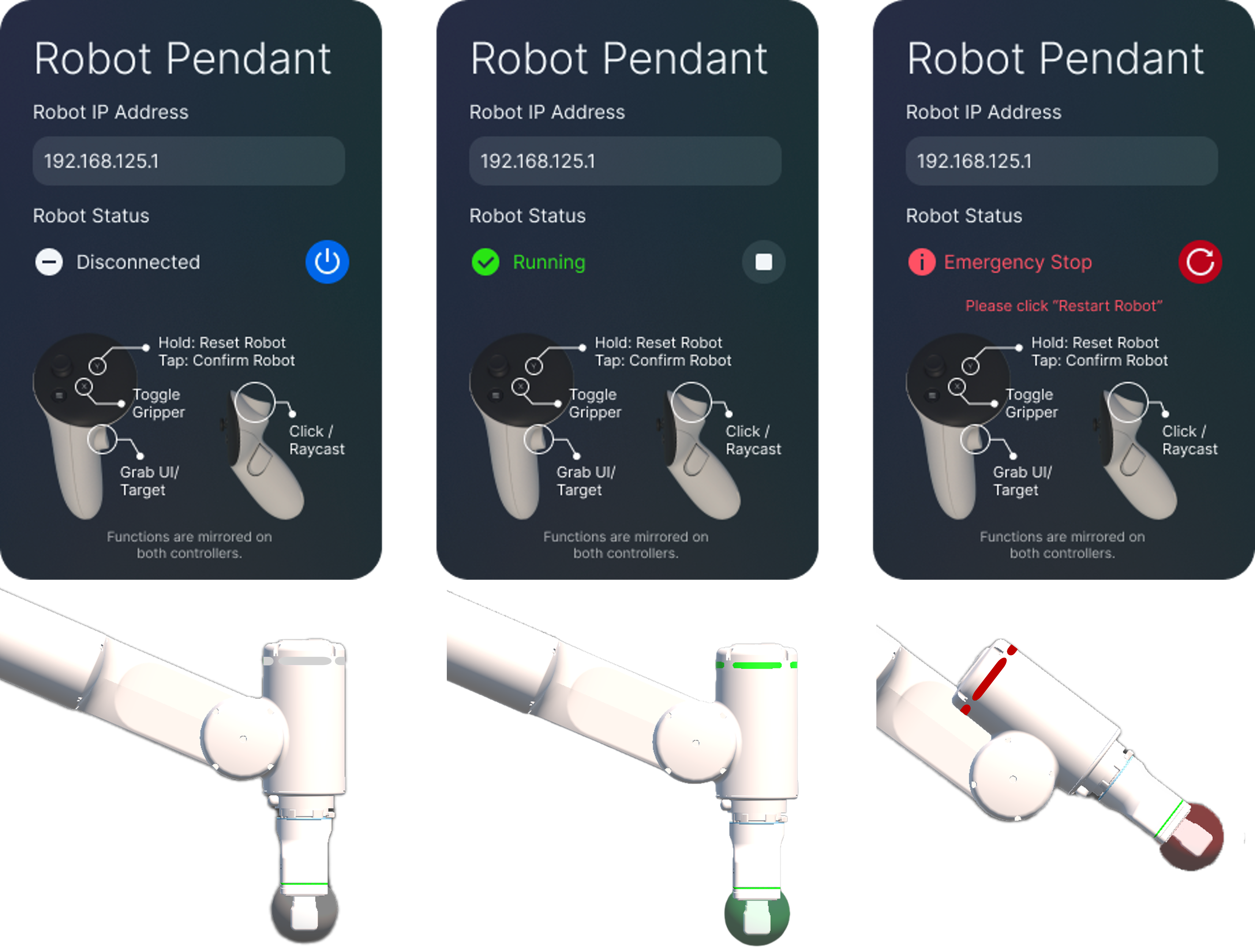}
\caption{AR User Interface Design. The interface integrates robot status visualization and interaction instructions. Low-frequency system functions (e.g., Connect, Restart) are accessible through the AR control panel, while high-frequency teleoperation commands (e.g., Target Control, Gripper Trigger) are mapped to physical controller buttons. The second row illustrates the color-coded TCP target used for system status indication: gray for disconnected, green for ready, and red for error.}
\label{fig:ui}
\end{figure}

To enhance user situational awareness, a visual status indicator is implemented on the \ac{TCP} target, as shown in Fig.~\ref{fig:ui}. Color-coded feedback communicates system status intuitively: gray for disconnected, green for ready, and red for error. This design enables rapid status recognition without requiring detailed textual feedback.

System functions are divided into low-frequency controls (e.g., connect, disconnect, restart) and high-frequency operations (e.g., target control, gripper trigger, pose confirmation/reset). A hybrid interaction model is adopted: low-frequency functions are integrated into the \ac{AR} \ac{UI}, while high-frequency commands are mapped to controller buttons. This design, as illustrated in Fig.~\ref{fig:ui}, minimizes workflow interruptions while maintaining efficient access to essential functions. Additionally, a configurable spatial anchor management system is developed to support flexible deployment in varying industrial layouts. This system provides accurate alignment of \ac{AR} content with the robot's physical workspace through flexible coordinate mapping mechanisms.

To streamline development and debugging, a custom debug interface was introduced, providing real-time system monitoring and parameter tuning without restarting the \ac{AR} session. Since each restart incurs a 30–50s delay, the ability to debug within a running session substantially reduces downtime and speeds up the iteration process.

\subsection{Spatial Point Cloud Rendering Module}

The Spatial Point Cloud Rendering Module aims to capture and render real-time 3D information of the robot workspace using depth sensors, enabling operators to perceive spatial context accurately from outside the safety enclosures in an \ac{AR} environment.

A comprehensive evaluation was conducted on mainstream depth sensors, including Intel RealSense D455, D435i, D405, and Microsoft Azure Kinect. Devices were assessed based on depth accuracy, working range, environmental robustness, and Unity compatibility. Testing showed that the D455 and Azure Kinect offered the best long-range performance ($>$1.5m), while the D435i performed well at mid-range (0.8–1.5m). Despite minor flicker, the D405 provided the highest detail resolution for close-range applications ($<$0.8m).

Given the system’s first-person \ac{AR} rendering strategy, the Intel RealSense D405 is selected. Its close-range precision aligns with the typical operating distance of end-effector-mounted vision and provides intuitive, real-time spatial awareness from the robot’s perspective. Mounted at an angle on the robot’s end-effector, the camera configuration minimizes occlusion from the gripper and expands the field of view, ensuring consistent visual access to the workspace during manipulation.

To improve the quality and stability of rendered point clouds, a multi-stage post-processing pipeline is implemented to mitigate noise, holes, and temporal jitter caused by sensor artifacts and environmental conditions. The pipeline includes:
\begin{itemize}
\item Threshold Filter: Removes data outside the 0–1m range to isolate the robot workspace.
\item Disparity Transform: Converts depth data to disparity space for more effective edge-preserving filtering.
\item Spatial Filter: Reduces spatial noise and fills small gaps (magnitude = 2, alpha = 0.5, delta = 20).
\item Temporal Filter: Stabilizes frames using temporal coherence (alpha = 0.4, delta = 20, persistence = 3).
\item Inverse Disparity Transform: Converts the filtered data back to depth format for Unity rendering.
\end{itemize}

\begin{figure}[tb]
  \centering
  \includegraphics[width=\linewidth]{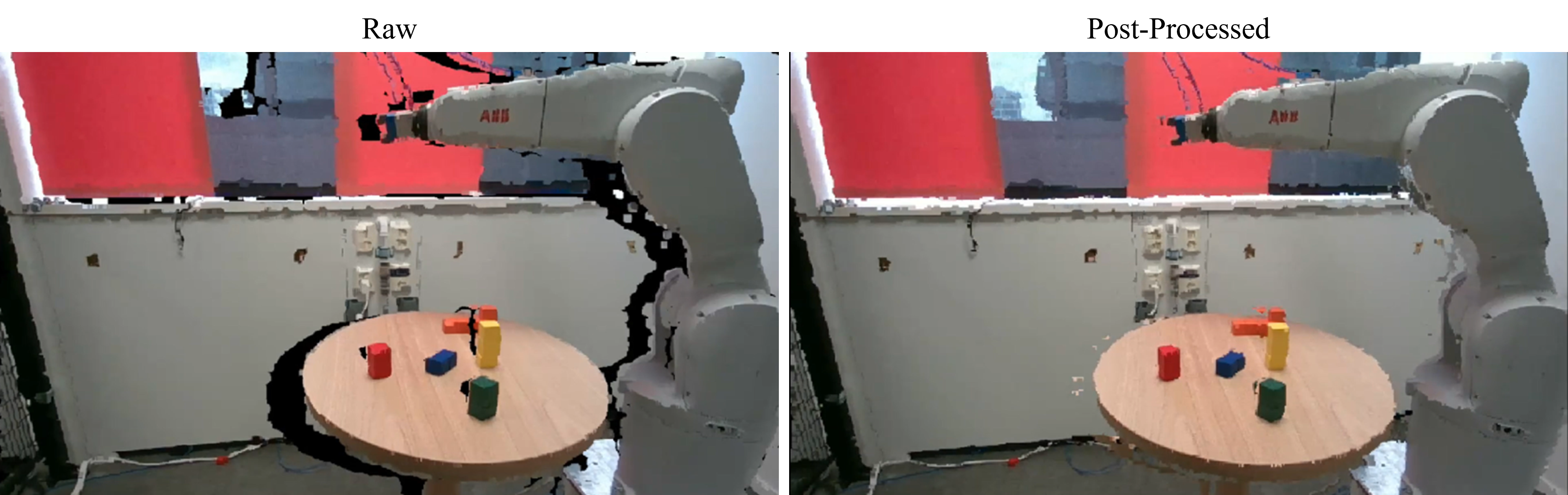}
  \caption{Comparison of Point Cloud Quality: Raw vs. Post-Processed.}
  \label{fig:postprocessing}
\end{figure}

As shown in Fig.~\ref{fig:postprocessing}, post-processing significantly improves spatial coherence and visual clarity compared to raw input, reducing jitter and preserving object geometry.

\subsection{Robot Monitoring Module}

The Robot Monitoring Module is developed to support reliable error detection and feedback in AR-based robot teleoperation, ensuring operational safety and system stability. It is built upon ABB’s \ac{RWS} protocol, which provides a standardized \ac{REST} \ac{API} for accessing real-time robot status and executing control commands from third-party applications.

The module employs a persistent WebSocket connection to subscribe to two primary data streams: robot status (RWS resources \textit{panel/ctrl-state}) and error logs (RWS resources \textit{elog/5} and \textit{elog/9}). As RWS does not offer a direct API for accessing detailed error descriptions, an indirect error retrieval mechanism is implemented. The system first captures the error sequence number (seqnum) via log subscriptions (elog), then issues secondary API queries to fetch detailed error log information associated with the sequence number. Five common error types are defined: Emergency Stop (90518), Speed Violation (90515), Proximity to Singularity (50456), Joint Out of Range (50027), Joint Load Too High (50055), along with a general Unknown category. This structured feedback system provides actionable information to users, improving operational awareness and enhancing safety. Moreover, gripper control is also managed through the RWS protocol.

To maintain WebSocket connection integrity, a legacy subscription cleanup mechanism is employed. Prior to establishing a new subscription, the system scans for existing WebSocket connections, extracts subscription IDs using regular expressions, and issues DELETE requests to the RWS subscription endpoint to remove obsolete connections. This approach prevents resource leaks and avoids instability caused by excessive lingering subscriptions.

The module’s error-handling capabilities were validated through fault injection testing in ABB RobotStudio. Each of the defined error types was individually triggered in a simulated environment. The system successfully detected and reported all events, confirming its ability to detect and manage various abnormal robot states in real time.

\section{User Study}\label{sec:exp}

A user study was conducted in this work with two primary objectives: (1) to evaluate the effectiveness and usability of the proposed AR-based robot teleoperation system, and (2) to examine the influence of real-time environmental perception on task accuracy, efficiency, and user confidence. 

\subsection{Experimental Setup}

\begin{figure}[tb]
\centering
\includegraphics[width=\linewidth]{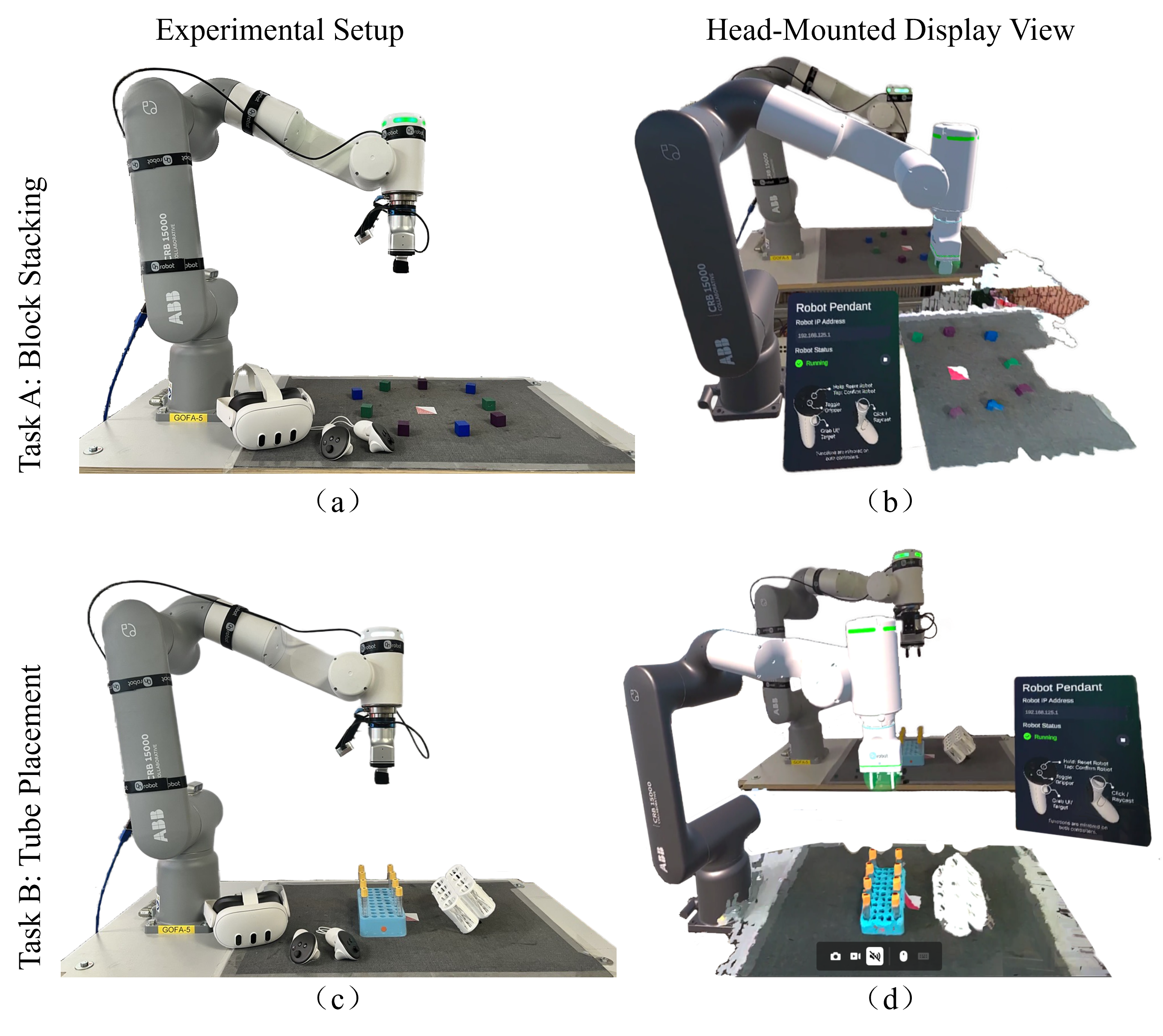}
\caption{Experimental setup for the user study. (a) Physical setup and (b) HMD first-person view for Task A: block stacking. (c) Physical setup and (d) HMD view for Task B: tube placement.}
\label{fig:setup}
\end{figure}

The experimental setup (Fig.~\ref{fig:setup}) comprises the following components:

\paragraph*{HMD} Meta Quest 3 is employed as the primary display interface, providing immersive \ac{AR} visualization with precise spatial tracking and interaction. As a video passthrough \ac{AR} HMD, it captures stereo camera feeds of the physical environment and renders them onto high-resolution screens, enabling real-time integration of virtual content. Compared to optical see-through systems such as Microsoft HoloLens 2, which overlay holograms onto transparent lenses, passthrough \ac{AR} offers enhanced visual fidelity, full-scene occlusion, and a wider field of view. Although passthrough introduces some latency, it is better suited for tasks requiring high spatial accuracy and realism, such as robot teleoperation.

\paragraph*{wireless controller} The Touch Plus controllers provide 6-DoF tracking and haptic feedback to support user interaction. Each controller integrates physical buttons, thumbsticks, and triggers.

\paragraph*{robot} The ABB GoFa 5 (CRB 15000) is a 6-axis collaborative robot with a 5kg payload capacity. In the user study, GoFa 5 serves as the execution unit, receiving motion commands and performing corresponding actions. While GoFa 5 is used as the primary demonstration platform, the proposed AR-based teleoperation system was also implemented and validated on the ABB IRB 1200 industrial robot, demonstrating cross-platform compatibility.

\paragraph*{depth camera} An Intel RealSense D405 depth camera is mounted on the robot’s end-effector, capturing real-time RGB-D images. It provides depth data and generates 3D point clouds for environmental perception.

\paragraph*{manipulated items} Two tasks were designed for the user study: one serving as a proof-of-concept in a controlled lab setting, and the other reflecting a realistic lab automation scenario. The block stacking task (Task A) requires users to teleoperate the robot to vertically stack colored wooden blocks within a predefined target zone marked by a red-and-white rectangular sticker. The tube placement task (Task B) involves teleoperating the robot to pick up transparent tubes with yellow caps, filled with simulated liquid, from a blue tube rack and insert them into designated slots on a white rack. For examples of both tasks, please refer to the accompanying video\footnote{\url{https://youtu.be/PiWwtgK0sSQ}}.

\paragraph*{first-person AR view} Fig.~\ref{fig:setup} illustrates the first-person view as seen through the \ac{HMD}. The background depicts the physical environment, while the foreground overlays virtual content. The augmented elements include the \ac{UI} panel, the virtual robot model, and real-time point cloud data.

\subsection{Experimental Procedure}

In line with research ethics, all participants provided informed consent prior to the study, acknowledging their understanding of the study’s purpose, procedures, data usage and storage. All data were anonymized.

The experimental tasks were designed to reflect typical industrial scenarios, with safety considerations requiring all operations to be conducted at least 1 meter from the robot workspace. This setup aligns with the core requirement of avoiding physical entry into the robot's workspace.

Participants were first given 10 minutes to familiarize themselves with the system, including both tasks: block stacking (Task A) and tube placement (Task B). During this practice period, participants explored the \ac{UI}, tested the controllers, anchored the virtual robot, and teleoperated the robot’s \ac{TCP}.

A within-subjects design was adopted, where each participant completed both tasks under two system conditions: with and without point cloud rendering. The control condition offered basic AR-based robot teleoperation as in \cite{van2024puppeteer, smith_augmented_2025}, while the experimental condition incorporated real-time point cloud visualization of the environment, rendered in front of the virtual robot. All other aspects, including \ac{UI} layout, interaction mechanisms, and control logic, remained consistent across both configurations.

To mitigate learning and fatigue effects, a balanced Latin square design was used. Five participants took part in the study, with three beginning in the point cloud condition and two starting in the control condition.

Under each system condition, participants completed two trials of Task A and two of Task B. Each trial began with system initialization and connection setup, including virtual robot positioning and establishing communication with the physical robot. For Task A (block stacking), participants had 3 minutes to pick and vertically stack 9 colored blocks within a target area. For Task B (tube placement), participants had 3 minutes to insert 8 transparent tubes from a blue rack into slots on a white rack. 

After completing the four trials in each system condition, participants completed the \ac{SUS} questionnaire to evaluate system usability. \ac{SUS} \cite{brooke_sus_1995, brooke_sus_2013, lewis_system_2018} is a widely used standardized assessment tool that measures usability across dimensions such as ease of use, complexity, learnability, and user confidence, using 10 standardized questions, each rated on a 5-point Likert scale (1 = strongly disagree to 5 = strongly agree).

Following completion of all trials and SUS assessments, participants engaged in a semi-structured interview to provide qualitative feedback. The interview design followed the five planes of user experience framework \cite{garrett_elements_2010}, covering:
\begin{itemize}
\item Strategy Layer: How likely would you be to use this system in your actual work? What are your thoughts on recommending it for industrial collaboration?
\item Scope Layer: How would you evaluate the overall system functionality? What features would you like to see improved or added?
\item Structure Layer: How would you describe your interaction experience with the system?
\item Skeleton Layer: How intuitive did you find the virtual robot control and spatial positioning? What was your experience with object grasping and operational precision?
\item Surface Layer: How would you rate the visual presentation and \ac{AR} rendering quality? What impact did the point cloud display have on your operation?
\end{itemize}

\subsection{Performance Metrics}

To assess the effectiveness of task execution in alignment with industrial robot application requirements, three core performance metrics were defined:

\paragraph{Maximum Item Completion Count ($N_{max}$)} Measures the highest number of items successfully manipulated within the time limit, specifically, the number of blocks stacked and tubes placed. This metric evaluates task efficiency and operational fluency.

\paragraph{Minor Error Count ($E_{minor}$)} Captures general operational errors, including failed grasps, object drops during manipulation, and imprecise placements. It reflects control accuracy and user proficiency.

\paragraph{Major Error Count ($E_{major}$)} Identifies critical failures, defined per task context. For block stacking: robot system faults and tower collapse; for tube placement: robot errors, object damage, and rack collapse. This metric assesses system safety and robustness.

Data were collected through direct observation by researchers, who recorded abnormal situations and operational challenges during each trial.

\section{Results and Discussion}\label{sec:results}

\subsection{User profiles and subjective evaluation}
\begin{figure}[tb]
\centering
\includegraphics[width=\linewidth]{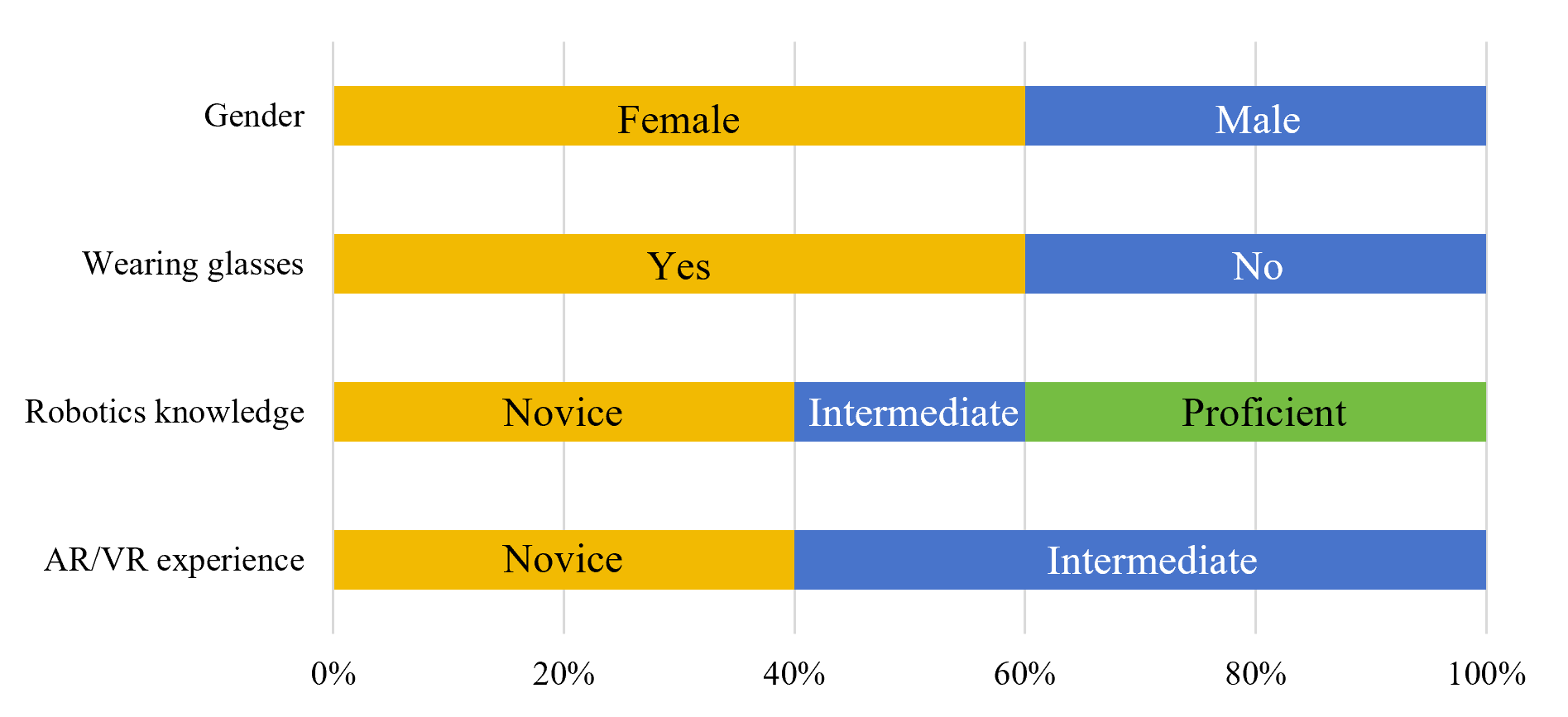}
\caption{Overview of the five participant profiles, detailing their AR/VR experience, robotics knowledge, gender, and eyewear usage.}
\label{fig:userprofile}
\end{figure}

Five participants, aged 25 to 33 with engineering backgrounds, took part in the user study. Their profiles are summarized in Fig.~\ref{fig:userprofile}, including information on \ac{AR}/\ac{VR} experience, robotics knowledge, gender, and eyewear usage. Experience levels were categorized into three groups: Novice (no prior experience), Intermediate (basic familiarity or occasional use), and Proficient (frequent use with advanced understanding). The participant pool showed balanced distribution across experience levels, gender, and eyewear use.

\begin{figure}[tb]
\centering
\includegraphics[width=\linewidth]{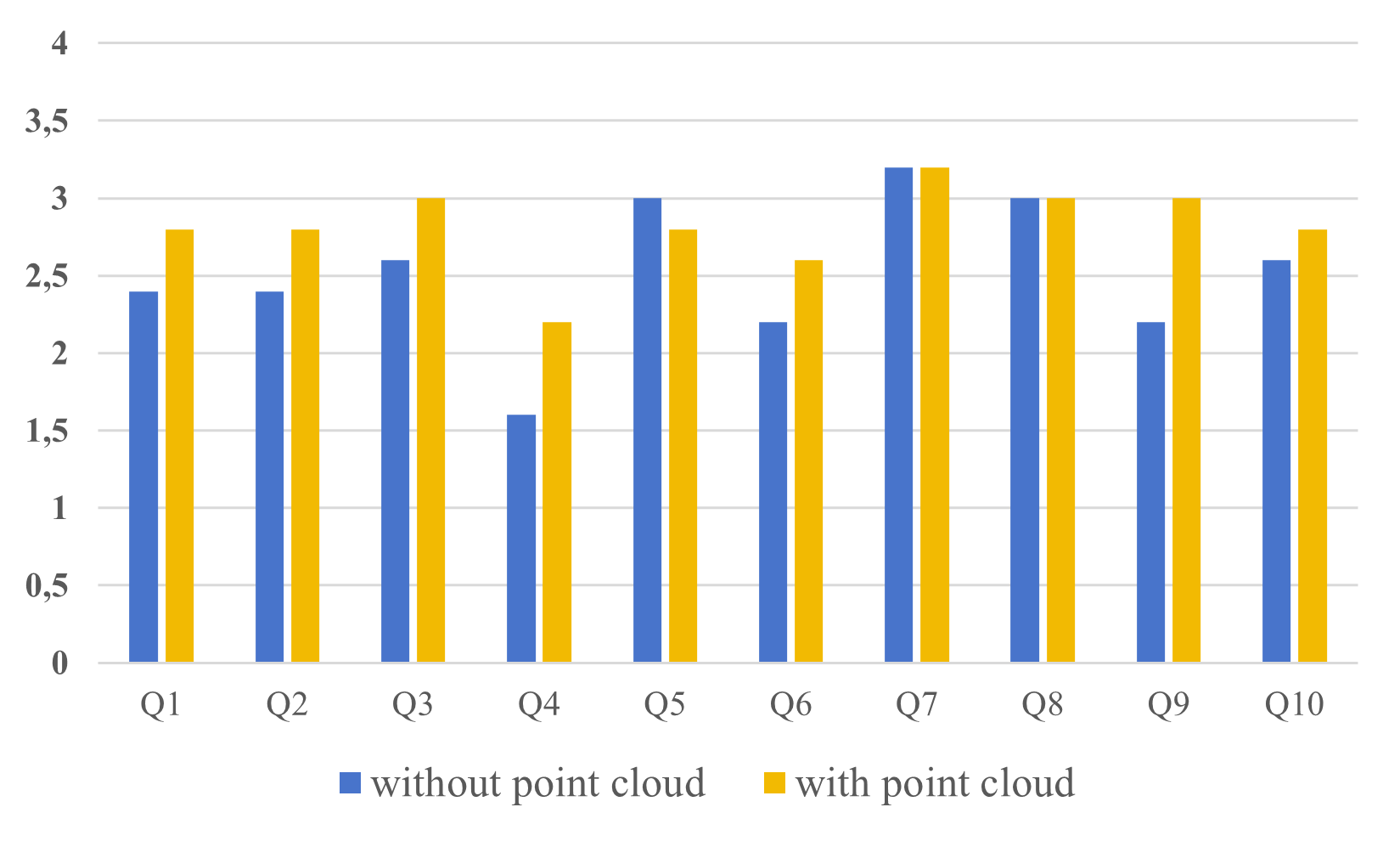}
\caption{Comparison of SUS responses across 10 standardized items under two conditions: with and without point cloud rendering. Each bar represents the average participant rating on a 5-point Likert scale.}
\label{fig:sus}
\end{figure}

The perceived usability of the system, evaluated using SUS, is illustrated in Fig.~\ref{fig:sus}. The chart compares user responses across 10 standardized items under two conditions: with and without point cloud rendering.

Responses were collected using a 5-point Likert scale. To normalize scores, odd-numbered items (positive statements) were adjusted by subtracting 1, and even-numbered items (negative statements) by subtracting from 5. The resulting scores for each item, ranging from 0 to 4, were averaged and plotted. The overall SUS score was computed by summing the normalized item scores and multiplying by 2.5, yielding a total score out of 100.

The system without point cloud received an overall SUS score of 63, categorized as ``OK" per the grading scale in \cite{brooke_sus_2013}, while the configuration with point cloud achieved 70.5, corresponding to a ``Good" usability rating. The results indicate that the point cloud-enhanced system was generally perceived as more usable across most items. An exception was Question 5 (“I found the various functions in this system were well integrated”), where negative feedback on point cloud use slightly impacted the score.

We assessed differences between system conditions using statistical tests selected according to data distribution. Normality of each dataset was assessed with the Shapiro–Wilk test ($\alpha$ = 0.05). Since both “system with point cloud” (p = 0.751) and “system without point cloud” (p = 0.112) met the normality assumption, we applied Welch’s t-test to compare the mean SUS scores. For all tests, the significance threshold was set at $\alpha$ = 0.05, and 95\% \ac{CI} was computed for the mean differences. Effect size was quantified using Cohen’s d, and post-hoc power analysis was conducted to assess the adequacy of the sample size. A valuable direction for future work would be to replicate this study with a larger sample size to achieve higher statistical power and provide more definitive conclusions.

The mean SUS score for “system with point cloud” was 70.5 (SD = 23.14), compared to 63.0 (SD = 17.54) for the “system without point cloud” condition. Welch’s t-test indicated that this difference was not statistically significant (t(6.87) = 0.578, p = 0.581, 95\% CI [–22.83, 37.83]). The effect size was small-to-moderate (d = 0.37). Post-hoc power analysis revealed very low statistical power (8\%) for detecting an effect of this magnitude with the current sample (five participants), suggesting that the non-significant result may be attributable to insufficient sample size rather than the absence of a true difference. A valuable direction for future work would be to replicate this study with a larger sample size, approximately 80 participants, to achieve higher statistical power and provide more definitive conclusion.

\subsection{Semi-structured interview} 

Semi-structured interviews were conducted to gather qualitative feedback from participants. Key findings and discussions are summarized below:

\paragraph{Intuitiveness of contactless interaction}
Participants found controller-based interaction intuitive and easy to learn. However, fine manipulation required further refinement. Some users reported excessive sensitivity during grasping near targets, suggesting the need for adaptive speed control.

\paragraph{AR interface design}
The \ac{AR} interface supported basic spatial awareness, while the auto-following \ac{UI} panel was frequently perceived as distracting. Limited visual cues led users to operate within the visible range, at times blocking the point cloud. Prolonged \ac{HMD} use caused discomfort, particularly for glasses wearers. Most participants found the virtual robot model unnecessary, as it occluded the real scene. Future designs may prioritize visualizing only the end-effector and gripper, omitting the full robot arm.

\paragraph{Error handling and collision awareness}
The current error handling, requiring full robot resets, interrupted workflow and reduced user confidence. Participants also noted insufficient detection of side and rear obstacles, highlighting the need for more comprehensive collision awareness beyond the current point cloud implementation.

\paragraph{Point cloud rendering}
Point cloud data significantly enhanced spatial alignment and object localization, especially for larger objects like blocks. Four out of five participants preferred the point cloud-enabled system. However, frequent switching between real-world observation and point cloud visualization increased cognitive load, slightly reducing task efficiency. Users also reported concerns about flickering, occlusions, and limited field of view. Precise alignment during object placement remained the most challenging task. While most users expressed confidence in using the system in industrial contexts, several emphasized that point cloud data should supplement rather than replace direct visual inspection, especially for high-precision operations.

\subsection{Objective evaluation}

\begin{figure}[tb]
\centering
\includegraphics[width=\linewidth]{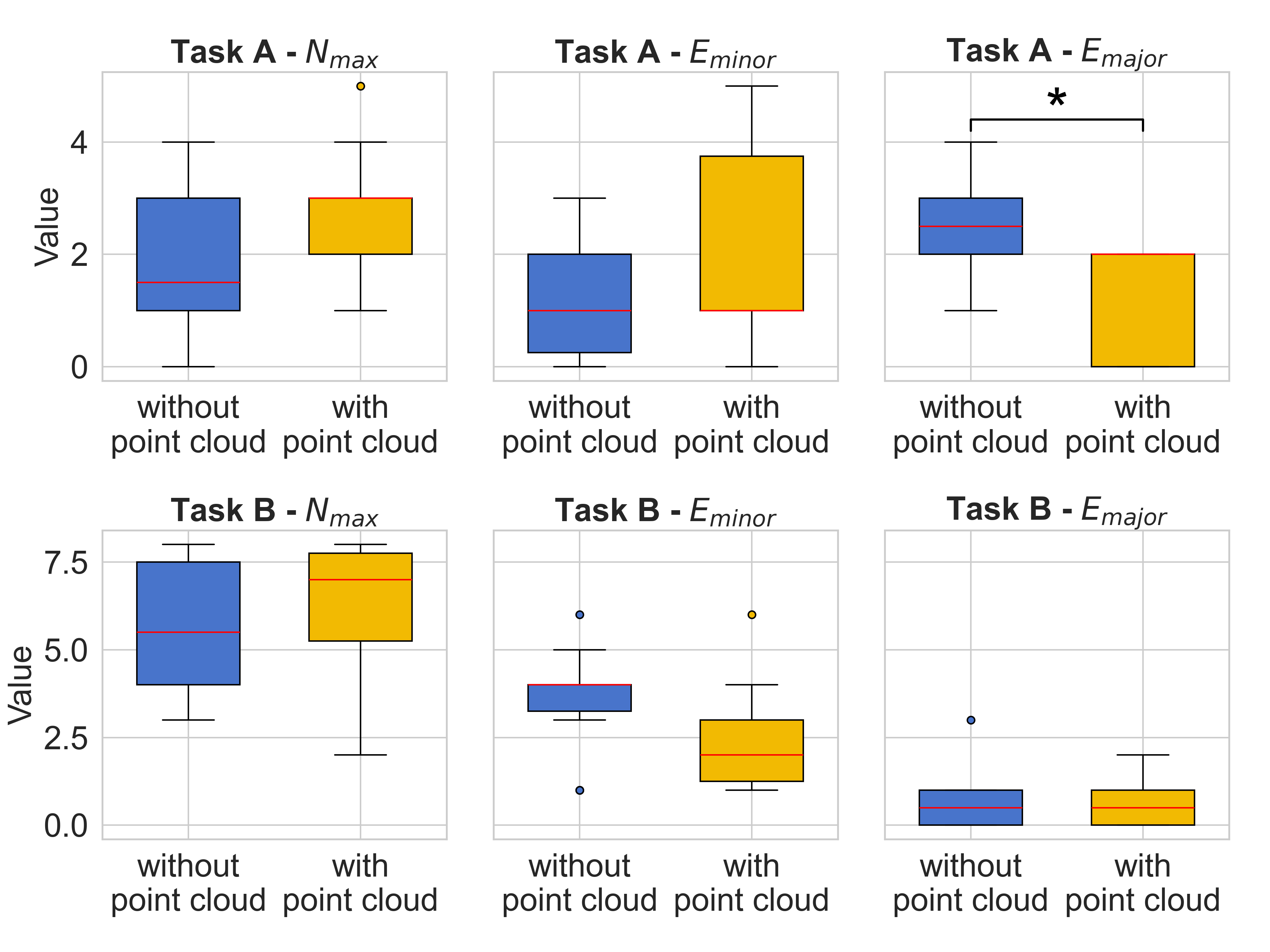}
\caption{Comparison of performance metrics with and without point cloud input across two tasks. Each subplot shows boxplots for 10 trials per condition. Task A and Task B results are organized in the first and second rows, respectively. Metrics include Maximum Item Completion Count ($N_{max}$), Minor Error Count ($E_{minor}$), and Major Error Count ($E_{major}$). Boxes represent medians, interquartile range, and data's minimum-maximum span. A line labeled with an asterisk connecting two groups indicates a significant difference between them (i.e., $p<0.05$ using Kruskal-Wallis test or t-test).}
\label{fig:boxplot}
\end{figure}

Fig.~\ref{fig:boxplot} presents a comparative analysis of task performance metrics under two system configurations: with and without point cloud rendering, across both Task A (block stacking) and Task B (tube placement).

In Task A, the inclusion of point cloud rendering led to a notable improvement in task efficiency and operational fluency: the mean $N_{max}$ increased from 1.9 to 2.7. Meanwhile, $E_{major}$ dropped significantly from 2.5 to 1.2, suggesting enhanced system safety and robustness, as fewer critical failures (e.g., tower collapses or system errors) were observed. For the $E_{major}$ performance metric, “system with point cloud” (median = 1.0) was significantly lower than the “system without point cloud” condition (median = 3.0). Because the normality assumption was violated (Shapiro–Wilk $p<0.05$), we used nonparametric tests. A Kruskal–Wallis test indicated a significant effect ($\chi^{2}(1)=5.35$, $p=0.021$), and a Mann–Whitney U test confirmed that the inclusion of point cloud rendering scored significantly lower than the “system without point cloud” condition (U = 21.0, p = 0.023, 95\% CI for median difference [–3.0, 0.0]). However, $E_{minor}$ rose from 1.3 to 2.0, reflecting a marginal increase in manipulation errors such as failed grasps or imprecise placements.

In Task B, the system with point cloud support showed a higher $N_{max}$ (6.3 vs. 5.5), again suggesting better task fluency. $E_{minor}$ decreased from 3.6 to 2.5, indicating improved control accuracy and user proficiency during tube handling. $E_{major}$ showed a slight reduction from 0.7 to 0.6, suggesting consistent system reliability with minimal critical errors (e.g., object damage or rack collapse).

Overall, the results indicate that point cloud input generally enhances task performance, increasing completion counts by 28\% and reducing critical errors by 33\%, especially in spatially constrained manipulation tasks. 

Minor error trends varied between tasks under point cloud conditions. In Task A, which demands precise alignment during grasping, close-range interactions led to unstable or degraded point cloud rendering, increasing minor errors. This highlights the need for improved rendering quality in close-up views. To address this, a proof-of-concept study was conducted using sensor fusion with two depth cameras. Preliminary results indicate that multi-sensor integration is a promising direction for enhancing spatial perception in AR-based teleoperation systems.

\section{Conclusion}\label{sec:conclusion}
This study introduces a modular AR-based teleoperation framework designed for flexible deployment across industrial robot platforms. The system architecture integrates spatial perception, intuitive interaction, and real-time feedback, enabling effective task execution without physical proximity. Experimental findings underscore the system’s practical viability and provide insights into \ac{UI} design, sensor integration, and safety considerations for AR-assisted robot control. Future work will explore scalability across broader industrial applications and integration with learning-based automation.

\bibliographystyle{IEEEtran}
\bibliography{main}
\end{document}